%% file: main.tex
\documentclass[10pt,twocolumn,letterpaper]{article}

\usepackage{wacv}      

\usepackage{graphicx}
\usepackage{booktabs}
\usepackage{multirow}
\usepackage{amsmath,amssymb}
\usepackage{listings}
\usepackage{pifont}
\usepackage[table]{xcolor}

\newcommand{\impr}[1]{$_{\color{red!80!black}{\textbf{-#1}}}$}
\newcommand{\degr}[1]{$_{\color{gray}{+#1}}$}

\lstdefinestyle{prologstyle}{
  basicstyle=\ttfamily\small,
  keywordstyle=\bfseries,
  numbers=none,
  frame=single,
  backgroundcolor=\color[gray]{0.95},
  breaklines=true,
  captionpos=b,
  xleftmargin=1em,
  xrightmargin=1em
}


\definecolor{wacvblue}{rgb}{0.21,0.49,0.74}
\usepackage[pagebackref,breaklinks,colorlinks,allcolors=wacvblue]{hyperref}
\usepackage{latexml}

\def\wacvPaperID{1731} 
\def\confName{WACV}
\def\confYear{2027}

\title{A Neuro-Symbolic Framework Combining Inductive and Deductive Reasoning for Autonomous Driving Planning}

\iflatexml
  \author{
    Hongyan Wei\\
    Clemson University, United States\\
    \texttt{hongyaw@clemson.edu}
    \and
    Wael AbdAlmageed\\
    Clemson University, United States\\
    \texttt{wabdalm@clemson.edu}
  }
\else
  \author{
    Hongyan Wei \qquad Wael AbdAlmageed\\
    Clemson University, United States\\
    {\tt\small hongyaw@clemson.edu \qquad wabdalm@clemson.edu}
  }
\fi

\begin{document}
\maketitle

\begin{abstract}
Existing end-to-end autonomous driving models rely heavily on purely data-driven inductive reasoning. This ``black-box'' nature limits interpretability
and makes safety-critical planning challenging
in complex, long-tail scenarios. To overcome this bottleneck, we propose a novel neuro-symbolic trajectory planning (NS-Plan) framework, which integrates rigorous deductive reasoning into end-to-end neural networks. Specifically, our framework utilizes a Large Language Model (LLM) to dynamically extract scene rules and employs an Answer Set Programming (ASP) solver for deterministic logical arbitration, generating logically consistent and traceable discrete driving decisions. To bridge the gap between discrete symbols and continuous trajectories, we introduce a decision-conditioned decoding mechanism that transforms high-level logical decisions into learnable embedding vectors, simultaneously constraining the planning query and the physical initial velocity of a differentiable Kinematic Bicycle Model (KBM). By combining KBM-generated baseline trajectories with neural residual corrections, our approach uses the KBM to guide trajectory generation while keeping the decision-making process traceable. On the nuScenes benchmark, our method (NS-Plan) comprehensively outperforms state-of-the-art baselines, reducing the average $L_2$ error by 18.3\% (to 0.49 m), the collision rate by 22.2\% (to 0.07\%), and improving the trajectory prediction consistency (TPC) by 13\% (to 0.47 m) compared to MomAD. More importantly, in the complex Turning-nuScenes subset, our method reduces the collision rate by 78.1\% (to 0.07\%) over the baseline.

\vspace{2mm}
\noindent\textbf{Keywords:} Autonomous Driving Planning, Deductive Reasoning, Answer Set Programming, Kinematic Constraints
\end{abstract}

\section{Introduction}
\label{sec:intro}
Vision-based end-to-end autonomous driving systems aim to directly map raw sensor inputs to the vehicle's planned trajectory using a unified neural network architecture. In recent years, end-to-end methods such as UniAD \cite{hu2023planning}, VAD \cite{jiang2023vad}, and SparseDrive \cite{sun2024sparsedrive} have unified perception, prediction, and planning modules under a differentiable framework for joint optimization, continuously improving performance in benchmarks such as nuScenes \cite{caesar2020nuscenes}. MomAD \cite{song2025momad} further introduces a trajectory momentum mechanism, achieving new breakthroughs in the temporal consistency of trajectories. However, this inductive reasoning paradigm, which relies entirely on massive amounts of training data, faces serious theoretical limitations: purely ``black box'' neural networks lack common-sense reasoning capabilities and explicit physical and logical constraints. When faced with complex long-tail scenarios with out-of-distribution (OOD) training, the system can easily generate planned trajectories that defy common sense or even pose dangers. Furthermore, due to the lack of traceability in the internal decision-making process, researchers find it difficult to effectively attribute errors when trajectory planning fails.

Existing methods attempt to mitigate these issues by directly regressing coordinate displacements, which often yields kinematically infeasible trajectories, or by employing post-processing filters and data augmentation \cite{sun2024sparsedrive, dauner2023parting, jiang2023vad, bansal2018chauffeurnet}. However, post-processing cannot inherently correct the neural generation process, and data augmentation struggles to generalize to unseen edge cases.

To address this issue, traditional solutions typically rely on purely symbolic systems or rule-based reasoning\cite{shalev2017formal}. However, manually defined rules are difficult to fully enumerate and struggle to handle complex, real-world traffic scenarios. Furthermore, symbolic reasoning yields discrete driving commands, making it a major challenge to convert them into continuous trajectories that satisfy kinematic constraints. Meanwhile, although methods that directly use Large Language Models (LLMs) possess strong semantic understanding, they fundamentally rely on data-driven inductive reasoning. This makes them easily generate ``hallucinations'' and lack strict logical deduction. Consequently, pure LLM-based planners cannot guarantee logical consistency and safety in their decisions. This is also why we introduced ASP \cite{gebser2012answer} for constraint satisfaction.

The decision-making processes of human drivers provide insights for solving these problems. As conceptualized in \textit{Thinking, Fast and Slow}~\cite{kahneman2011thinking}, human cognition operates through both System 1 (Fast) and System 2 (Slow). Humans complement the fast, instinctive pattern matching of System 1 with the deliberate, rule-based logical reasoning of System 2 for driving decision-making.

Inspired by this, we propose a novel end-to-end neuro-symbolic framework for autonomous driving trajectory planning (NS-Plan), combining deductive reasoning from classical artificial intelligence with inductive reasoning from modern deep learning to achieve both powerful feature representation and rigorous interpretability. To realize this, we must overcome two key challenges: (1) bridging the semantic gap between symbolic discrete decisions and continuous trajectory generation, and (2) ensuring the planned trajectory possesses explicit physical semantics to accurately comply with logical velocity constraints.

To address these challenges, our main contributions are as follows:

\noindent\textbf{Neuro-Symbolic Planning (NS-Plan) Framework.} We present a novel paradigm that transforms high-level ASP \cite{gebser2012answer} symbolic decisions into learnable embedding vectors, directly guiding the trajectory decoding process rather than acting as a passive post-processing filter.

\noindent\textbf{Physics-Based Trajectory with Residuals.} We replace traditional coordinate regression with a differentiable Kinematic Bicycle Model (KBM) \cite{polack2017kinematic} to yield a dynamically feasible basic trajectory. We also add neural spatiotemporal residuals to compensate for basic physical trajectory. This design provides explicit, semantically consistent access points for symbolic speed constraints.

\noindent\textbf{LLM-ASP Logical Reasoning Engine.} We use a Large Language Model as a knowledge base to extract driving rules and use Clingo as an ASP \cite{gebser2012answer} solver to do logical reasoning to get driving decisions. This architecture enables safer, traceable, and discrete driving decisions in diverse long-tailed scenarios without relying on training data coverage.

\section{Related Work}
\label{sec:related}

\subsection{End-to-End Autonomous Driving Planning}
End-to-end autonomous driving has shifted from simple imitation learning \cite{chen2020learning,wu2022trajectory} to Transformer-based multi-task frameworks \cite{prakash2021multi,chitta2022transfuser,li2022bevformer,hu2022stp3,renz2022plant}. Recent works like UniAD \cite{hu2023planning} jointly optimize perception and planning. VAD \cite{jiang2023vad} uses vectorized representations for efficiency, SparseDrive \cite{sun2024sparsedrive} introduces a sparse perception architecture, and MomAD \cite{song2025momad} applies trajectory momentum for better temporal consistency. However, these purely data-driven methods act as uninterpretable black boxes, mapping perception to trajectories inductively. Their performance heavily relies on training data coverage, easily failing in rare, out-of-distribution (OOD) scenarios where explicit logical reasoning based on traffic rules is required.

\subsection{LLM-Based Planning}
Large language models (LLMs) bring strong common-sense and zero-shot reasoning to autonomous driving \cite{yang2023llm4drive,fu2024drivelikehuman}. Methods like DriveGPT \cite{huang2024drivegpt}, DriveGPT4 \cite{guo2024drivegpt4}, and DriveMLM \cite{wen2023drivemlm} use LLMs to directly output driving actions. DriveLM \cite{sima2024drivelm} builds a question-answering reasoning chain relying on visual instruction tuning \cite{liu2024visual}, while DriveVLM \cite{tian2024drivevlm} proposes a chain-of-thought pipeline with a hybrid slow-fast system. Despite their semantic understanding, LLMs struggle with formal logical consistency, sometimes outputting contradictory decisions. Moreover, converting their natural language outputs into continuous, kinematically feasible physical trajectories remains a major challenge.

\subsection{Deductive Reasoning and Symbolic Methods}
Early symbolic planning (e.g., expert systems \cite{shalev2017formal}) is logically strict but struggles to scale to real-world traffic semantics. Recent neural-symbolic methods often treat reasoning and trajectory generation sequentially—using symbolic rules either to judge neural features or to post-process neural outputs. This serial post-optimization does not affect the neural decoding process itself, thus failing to inherently guarantee safe planning. For instance, Sharifi et al. \cite{sharifi2023safe} combine logic with reinforcement learning for safe highway driving. General neural-symbolic frameworks like DeepProbLog \cite{manhaeve2018deepproblog} embed rules into training, but are rarely applied to complex 3D real-time planning. NeurASP~\cite{yang2020neurasp} treats neural network outputs as probabilities for ASP facts and trains end-to-end through the inference process. In contrast, we use Clingo to make deterministic decisions that guide a downstream differentiable trajectory decoder. This allows the symbolic reasoning component to provide strict logical constraints on the decision, while the neural network optimizes the final trajectory.

\subsection{Physics-Constrained Trajectory Planning}
Direct coordinate regression often ignores the physical boundaries of real-world vehicle motion. Physics-based models, such as the Kinematic Bicycle Model (KBM), address this by providing strict kinematic constraints and are common in traditional planning (e.g., MPC). Recently, models like KING \cite{hanselmann2022king} and PlanTF \cite{cheng2024rethinking} have used KBMs within neural networks to generate trajectory candidates for scoring. However, their control variables are not optimized end-to-end. Rule-based models like IDM \cite{treiber2000congested} offer precise speed control but fail in complex interactive scenarios. Cui et al.~\cite{cui2020deep} embed a kinematic bicycle layer to predict surrounding vehicles' future motion, demonstrating the benefit of physical constraints in trajectory prediction. However, their model targets other-agent prediction with no symbolic decision input. In contrast, we embed a differentiable KBM into the ego planner and directly couple it with ASP-derived logical decisions, enabling end-to-end optimization that translates discrete driving commands into kinematically feasible ego trajectories.

\subsection{Neuro-Symbolic Embodied Planning}
NS-Plan aligns with recent works that connect high-level reasoning (using symbols or language) to continuous physical actions. For embodied navigation, EvolveNav~\cite{lin2025evolvenav} uses LLM step-by-step reasoning to make clear navigational decisions. This shares our goal of making high-level reasoning easy to trace. Other constraint-aware methods~\cite{chen2025canav} extend this idea to zero-shot planning in continuous spaces. 
For robotic manipulation, RoBridge~\cite{zhang2025robridge} uses a two-level design to connect high-level planning and low-level execution. This is very similar to our dual-path design, which maps ASP decisions into both a spatial path and a kinematic speed profile. Additionally, A0~\cite{xu2025a0} uses Vision-Language-Action models to turn high-level reasoning into physical robot actions. This solves the same gap between semantic intent and physical control that motivates our decision-conditioned KBM decoder.
These previous works mainly focus on indoor navigation and manipulation. In contrast, NS-Plan focuses on autonomous driving. Driving requires extra constraints, such as traffic rules, physical limits, and real-time safety. Instead of using pure LLM reasoning, we use ASP. This provides deterministic and formally verifiable decisions, which are much better for safety-critical tasks.

\section{Method}
\label{sec:method}

\begin{figure*}[t]
  \centering
  \includegraphics[width=0.97\linewidth]{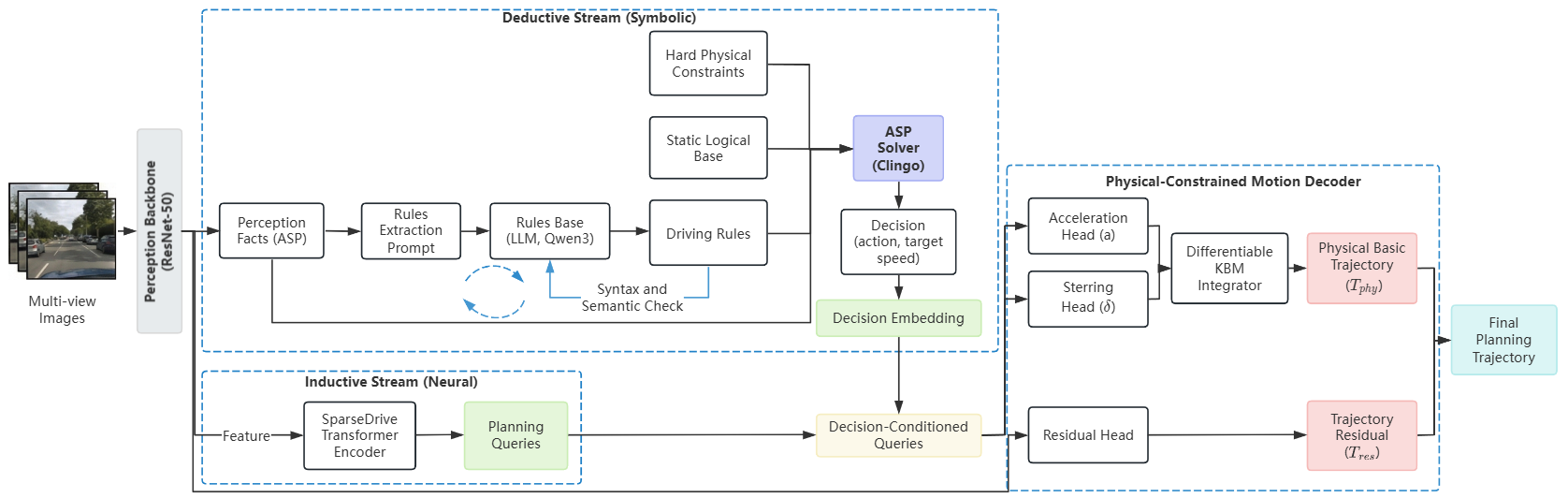}
  \caption{Overall framework. The inductive stream (neural) and 
  deductive stream (symbolic) run in parallel. The ASP engine 
  produces discrete driving decisions which 
  are conditioned into the Physics-Constrained Motion Decoder 
  via dual paths: Path 1 modulates planning 
  queries $\mathbf{Q}'_\text{plan} = \mathbf{Q}_\text{plan} 
  \oplus \mathbf{d}$ (spatial shape), and Path 2 biases the KBM initial velocity $v_0' = v_0 + b_v$ (speed 
  profile).}
  \label{fig:framework}
\end{figure*}

\subsection{Overall Architecture}
\label{sec:method:overview}

As illustrated in Fig.~\ref{fig:framework}, the proposed NS-Plan framework consists of three core modules: (1) Perception Fact Extraction, (2) Dynamic Deductive Reasoning Engine, and (3) Decision-Conditioned Physical Residual Planner.

\subsection{Deductive Reasoning Engine}
\label{sec:method:kbm}

To effectively combine inductive perception with deductive reasoning, a deterministic logical reasoning engine is required. Declarative Programming is an ideal paradigm for this task: rather than instructing the computer on \textit{how} to solve a problem step-by-step, we simply describe the facts and constraints, and the engine automatically finds the valid answers. We use Answer Set Programming (ASP) due to its exceptional capability in multi-conditional reasoning and complex conflict resolution. This perfectly aligns with how humans make driving decisions (e.g., stopping for a pedestrian even when the traffic light is green). Specifically, we use the widely-adopted tool Clingo \cite{gebser2012answer} as our ASP solver.

\begin{figure}[t]
  \centering
  \includegraphics[width=0.97\linewidth]{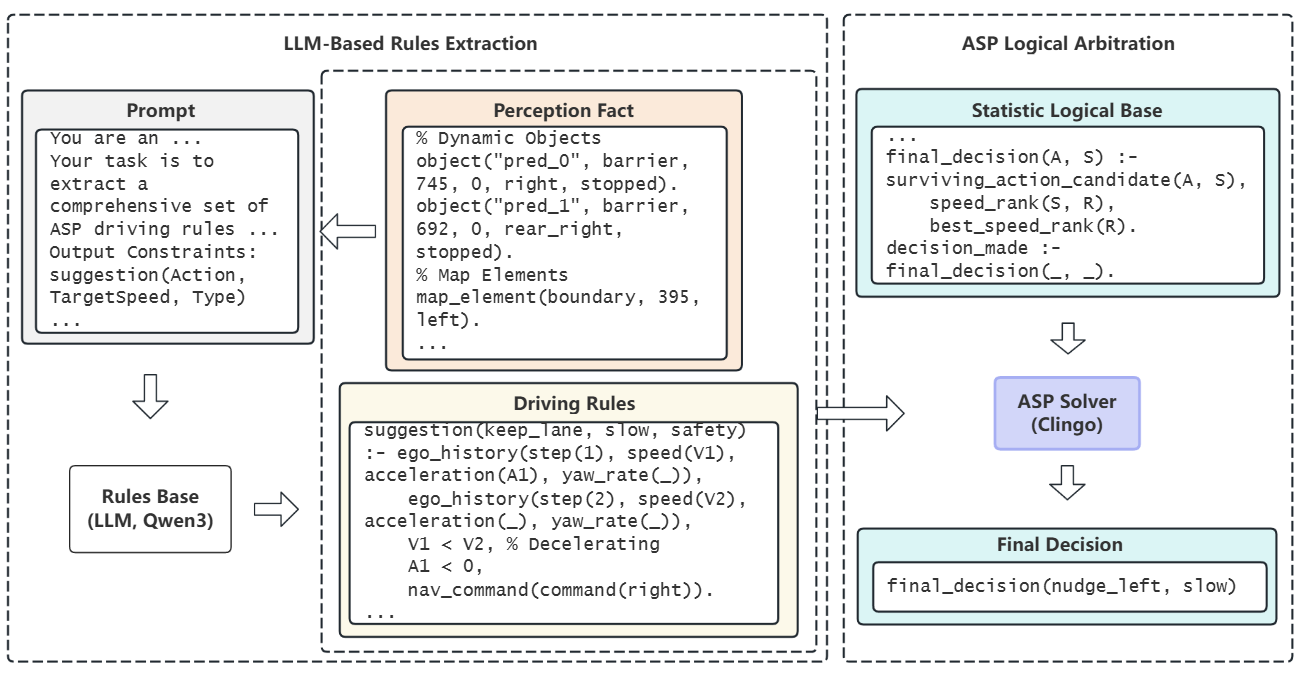} 
  
  \caption{Detailed pipeline of the deductive reasoning engine. The LLM extracts semantic rules from perception facts via prompts, which are then combined with a static logical base and evaluated by the ASP solver (Clingo) to deduce final deterministic driving decisions.}
  \label{fig:deductive_reasoning}
\end{figure}

To seamlessly incorporate these human traffic rules and common sense of physics into our system, our deployed ASP engine operates in two sequential stages as illustrated in Fig.~\ref{fig:deductive_reasoning}: dynamic rule extraction via an LLM, followed by deterministic arbitration using the Clingo solver.

\noindent\textbf{Perception to Fact Mapping.}
The output of the underlying perception module is structured into strict ASP predicates. For example, dynamic obstacles are represented as predicates with precise parameters: \emph{object(id, category, distance, speed, heading, relative\_pos, attribute, ttc)}. We reuse the ego-status prediction module from MomAD~\cite{song2025momad} to obtain predicted ego states, which, together with their history, are converted into symbolic facts. TTC is also computed using the predicted ego velocity.

\noindent\textbf{LLM-Based Rules Extraction.}
Hand-crafted rules often struggle to encompass the extensive long-tail of real-world traffic scenarios. To overcome this limitation, we design a structured System Prompt (see ``Prompt'' in Fig.~\ref{fig:deductive_reasoning}) that defines the LLM's role as an ``autonomous driving rule and ASP programming expert.'' Conditioned on the extracted semantic facts, the LLM dynamically generates 3 to 6 ASP logic suggestions tailored to the specific context of the current frame. This generation process is governed by expert prior knowledge and rigorous syntactic constraints (e.g., predicate formatting and variable safety). To ensure seamless logical arbitration, the output is strictly standardized as \emph{suggestion(Action, TargetSpeed, Type)}. Here, \emph{Type} characterizes the underlying rationale of the generated rule, such as an \emph{emergency} intervention, a \emph{safety} precaution, or \emph{legal} compliance. The precise definitions and boundaries of the remaining driving state variables (\emph{Action} and \emph{TargetSpeed}) are strictly constrained by a pre-defined symbolic vocabulary, as detailed below.

\noindent\textbf{Symbolic Vocabulary Design.}
Mapping high-level logical reasoning to specific, continuous physical trajectories requires clear and explicit driving intentions. To ensure deterministic logical arbitration and eliminate semantic ambiguity, we carefully construct a compact, mutually exclusive symbolic vocabulary tailored for complex urban driving. This vocabulary defines the atomic physical actions of the ego vehicle. Our task is short-term trajectory planning (3 to 6 seconds), not global point-to-point navigation. These symbols are the final output of the LLM-ASP reasoning and serve as direct inputs for the physical control of the vehicle (upstream of the KBM). Rather than allowing free-form language generation, the LLM's output is strictly constrained to this pre-defined semantic space. This vocabulary is systematically distilled from expert driving knowledge to maximize expressiveness while ensuring zero redundancy or semantic overlap. Specifically, the vocabulary encompasses:

\begin{itemize}
    \item[(1)] Action (lateral, 9): \texttt{keep\_lane}, \texttt{change\_lane\_left}, \texttt{change\_lane\_right}, 
               \texttt{turn\_left}, \texttt{turn\_right}, \texttt{nudge\_left}, \texttt{nudge\_right}, 
               \texttt{yield}, and \texttt{emergency\_stop}.
    \item[(2)] Target Speed (longitudinal, 6): \texttt{current}, \texttt{zero}, \texttt{creep}, 
               \texttt{slow}, \texttt{normal}, and \texttt{fast}. Here, the \texttt{current} category refers to the
model-predicted ego velocity rather than the
ground-truth ego velocity.
    \item[(3)] Navigation Commands (3): \texttt{left}, \texttt{right}, and \texttt{straight}. These are separately provided inputs for short-term trajectory planning, consistent with MomAD~\cite{song2025momad}.
\end{itemize}

By restricting the LLM outputs to the predefined
symbolic vocabulary, we standardize the predicates
as \emph{suggestion(Action, TargetSpeed, Type)},
where Type specifies the rule category. This ensures a consistent format for downstream ASP reasoning. Navigation commands are provided separately and are not generated by the LLM.

To mitigate the inherent hallucination issues of LLMs, we introduce a syntax and semantic check mechanism (as shown in Fig.~\ref{fig:framework}). If the generated rules contain logical contradictions or formatting errors, the prompt and the specific error messages are fed back to the LLM for re-generation until the output is completely correct. This iterative verification ensures valid inputs for the downstream ASP solver, thereby allowing only logically rigorous and physically safe rules are executed.

\noindent\textbf{Conflict Resolution.}
To resolve conflicting rules and obtain a deterministic decision, the dynamic rules are fed into the Clingo solver along with predefined physical safety axioms (as shown in Fig.~\ref{fig:deductive_reasoning}). To resolve rule conflicts, we design a strict priority mapping: emergency avoidance has the highest priority, safety precautions hold medium priority, and compliance and efficiency objectives hold the lowest priority. This design does not rely on arbitrary hyperparameters requiring network optimization; instead, it represents a rigorous formal encoding of real-world traffic rules. By evaluating facts and filtering conflicting actions through this hierarchical logic, the solver outputs the globally optimal decision for the current frame: final\_decision(Action, TargetSpeed).

\subsection{Decision-Conditioned Trajectory Planning}

After obtaining safe discrete decision symbols, the core challenge is how to transform them into continuous control trajectories. To address this, we designed a physical residual programming module for decision conditionalization.

\noindent\textbf{Perception Backbone and Planning Queries.}
The underlying perception module in this paper adopts the sparse perception architecture of SparseDrive \cite{sun2024sparsedrive}, using ResNet-50 \cite{he2016deep} as the visual backbone network to extract 3D scene and map features. After spatiotemporal interaction, the perception network outputs a multimodal planning query, which is formally defined as:
\begin{equation}
  \mathbf{Q}_{\text{plan}} \in \mathbb{R}^{B \times 1 \times M \times D},
  \label{eq:query}
\end{equation}
where $B$ represents the batch size, $M=18$ represents the number of planning modes (6 trajectory candidates $\times$ 3 navigation commands), and $D=256$ represents the feature dimension. All subsequent contributing modules operate on $\mathbf{Q}_{\text{plan}}$, without altering the core perception backbone.

\noindent\textbf{Dual-Path Decision Embedding.}
We first map discrete Actions and Speeds into high-dimensional decision feature vectors through an embedding layer. These features are then used to conditionally decode the trajectory via a dual-path mechanism: the first path injects the decision features as a semantic offset into the plan query to constrain the spatial shape of the trajectory; the second path learns a velocity bias parameter, which is directly used to correct the initial velocity of the physical model, thereby forcing the network to match the target velocity indicated by the underlying logic.

\noindent\textbf{Differentiable Kinematic Bicycle Model.}
Existing end-to-end methods directly regress coordinate displacements, lacking physical constraints. We explicitly introduce a differentiable kinematic model (KBM) into the network decoder. 

\begin{figure}[t]
  \centering
  \includegraphics[width=0.7\linewidth]{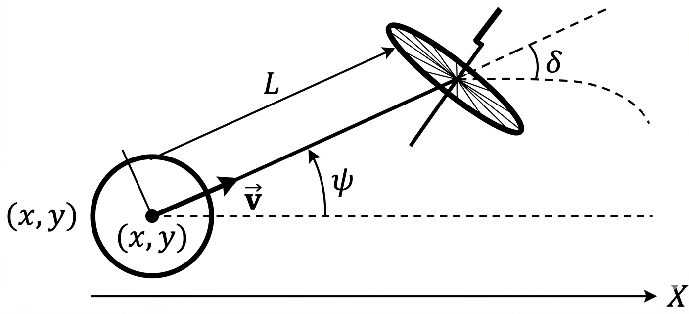}
  \caption{Illustration of the Kinematic Bicycle Model (KBM).}
  \label{fig:kbm}
\end{figure}

Given the network-predicted acceleration $a$ and front wheel steering angle $\delta$, the vehicle's state at time step $t$ is physically integrated using the second-order Runge-Kutta method (RK2), as depicted in Fig.~\ref{fig:kbm}:

\begin{equation}
  \dot{x} = v\cos\psi,\quad
  \dot{y} = v\sin\psi,\quad
  \dot{v} = a,\quad
  \dot{\psi} = \frac{v\tan\delta}{L},
  \label{eq:kbm_ode}
\end{equation}

where $x$ and $y$ denote the 2D spatial coordinates of the vehicle, $v$ is the velocity, $\psi$ represents the heading (yaw) angle, $a$ is the acceleration, $\delta$ is the front wheel steering angle, and $L$ represents the vehicle's wheelbase. The dot notation (e.g., $\dot{x}$) indicates the time derivative of the respective state variables. The KBM initial velocity $v_0$ is obtained from the predicted ego state rather than ground-truth velocity. The model integrates the initial velocity after ``velocity offset'' correction to generate a baseline trajectory $\boldsymbol{\tau}_{\mathrm{physics}}$ that conforms to vehicle dynamics constraints.

\noindent\textbf{Physics + Residual Prediction.}
Although the differentiable KBM provides explicit kinematic priors, it relies on two main control parameters: acceleration $a$ and front wheel steering angle $\delta$. In real-world driving, vehicle control is more complex, involving vehicle inertia, tire slip, and highly dynamic multi-agent interactions. Therefore, we introduce a spatiotemporal neural residual $\boldsymbol{\tau}_{\mathrm{residual}}$ as a complementary refinement mechanism. This design utilizes the basic physical trajectory as a reliable safety foundation, while using the neural residual to dynamically refine and compensate for complex boundary conditions, ultimately producing trajectories that are both kinematically feasible and highly adaptive to complex environments. The final planned trajectory $\boldsymbol{\tau}_{\mathrm{final}}$ is formulated as:
\begin{equation}
\boldsymbol{\tau}_{\mathrm{final}} = \boldsymbol{\tau}_{\mathrm{physics}} + \lambda \cdot \tanh(\boldsymbol{\tau}_{\mathrm{residual}}).
\label{eq:traj_final}
\end{equation}
In Eq.~(\ref{eq:traj_final}), the $\tanh$ function is employed to bound the residual magnitude, preventing non-physical corrections, while $\lambda$ acts as a scaling factor to balance the analytical baseline $\boldsymbol{\tau}_{\mathrm{physics}}$ and the learned refinement. This residual design ensures that the final trajectory $\boldsymbol{\tau}_{\mathrm{final}}$ remains physically plausible while being adaptive to dynamic environments.

\noindent\textbf{Loss Functions.}
We train the NS-Plan framework end-to-end using a total loss $\mathcal{L}$, which combines the basic motion planning loss with three specific physical and logical losses:
\begin{equation}
    \mathcal{L} = \mathcal{L}_{\mathrm{motion}} + w_1\mathcal{L}_{\mathrm{res}} + w_2\mathcal{L}_{\mathrm{smooth}} + w_3\mathcal{L}_{\mathrm{aux}},
\end{equation}
where $\mathcal{L}_{\mathrm{motion}}$ includes the standard baseline losses (e.g., coordinate regression and classification), and $w_1, w_2, w_3$ are weights to balance the terms. 

To follow physical rules, we introduce a residual loss $\mathcal{L}_{\mathrm{res}}$:
\begin{equation}
    \mathcal{L}_{\mathrm{res}} = \left(\alpha \cdot \mathcal{L}_1(\boldsymbol{\tau}_{\mathrm{res}}^x, \mathbf{0}) + \mathcal{L}_1(\boldsymbol{\tau}_{\mathrm{res}}^y, \mathbf{0})\right) \odot \mathbf{M}_{\mathrm{gt}},
\end{equation}
where $\alpha$ is a scaling factor designed to heavily penalize unnatural lateral sideslip (empirically set to $10$), $\mathcal{L}_1$ is the L1 distance, while $\boldsymbol{\tau}_{\mathrm{res}}^x$ and $\boldsymbol{\tau}_{\mathrm{res}}^y$ are the predicted residuals along the lateral (X-axis) and longitudinal (Y-axis) directions. $\mathbf{M}_{\mathrm{gt}}$ is a validity mask to ignore invalid timesteps.

To keep the vehicle's motion stable, we add a control smoothing loss $\mathcal{L}_{\mathrm{smooth}}$:
\begin{equation}
    \mathcal{L}_{\mathrm{smooth}} = \sum_{t=1}^{T-1} \left(|a_t - a_{t-1}| + |\delta_t - \delta_{t-1}|\right),
\end{equation}
where $T$ is the total number of future timesteps. The variables $a_t$ and $\delta_t$ are the predicted acceleration and steering angle at step $t$. By minimizing changes between consecutive frames, this loss ensures smooth physical control. 

Finally, the auxiliary loss $\mathcal{L}_{\mathrm{aux}}$ uses cross-entropy to align the pooled neural features with the ASP decisions. This keeps the neural and symbolic parts of our model strictly consistent.

\section{Experiments}
\label{sec:exp}

\subsection{Experimental Setup}

\begin{table*}[t]
\centering
\caption{Planning results on the nuScenes validation dataset. $*$ denotes results reproduced with the official checkpoint. Following SparseDrive~\cite{sun2024sparsedrive} and MomAD~\cite{song2025momad}, we use predicted rather than ground-truth ego-status for fair comparison. \textcolor{red!80!black}{Red} indicates the improvement. Our FPS is measured on H200, excluding LLM rule extraction; direct comparison is for 
reference due to hardware differences. NS-Plan results are reported as mean\,$\pm$\,std over 3 independent training runs with different seeds.}
\label{tab:main_results}
\setlength{\tabcolsep}{4pt} 
\resizebox{\textwidth}{!}{
\begin{tabular}{@{} l c c c c c >{\columncolor[gray]{0.92}}c c c c >{\columncolor[gray]{0.92}}c c c c >{\columncolor[gray]{0.92}}c c @{}}
\toprule
\multirow{2}{*}{Method} & \multirow{2}{*}{Input} & \multirow{2}{*}{Backbone} & \multicolumn{4}{c}{$L_2$ (m) $\downarrow$} & \multicolumn{4}{c}{Col. Rate (\%) $\downarrow$} & \multicolumn{4}{c}{TPC (m) $\downarrow$} & \multirow{2}{*}{FPS $\uparrow$} \\
\cmidrule(lr){4-7} \cmidrule(lr){8-11} \cmidrule(lr){12-15}
& & & 1s & 2s & 3s & \multicolumn{1}{c}{Avg.} & 1s & 2s & 3s & \multicolumn{1}{c}{Avg.} & 1s & 2s & 3s & \multicolumn{1}{c}{Avg.} & \\
\midrule
UniAD \cite{hu2023planning} & Camera & ResNet101 & 0.45 & 0.70 & 1.04 & 0.73 & 0.62 & 0.58 & 0.63 & 0.61 & 0.41 & 0.68 & 0.97 & 0.68 & 1.8 (A100) \\
VAD \cite{jiang2023vad} & Camera & ResNet50 & 0.41 & 0.70 & 1.05 & 0.72 & 0.03 & 0.19 & 0.43 & 0.21 & 0.36 & 0.66 & 0.91 & 0.64 & - \\
SparseDrive$^*$ \cite{sun2024sparsedrive} & Camera & ResNet50 & 0.29 & 0.58 & 0.96 & 0.61 & \textbf{0.01} & 0.05 & 0.18 & 0.08 & 0.30 & 0.57 & 0.85 & 0.57 & 9.0 (RTX4090) \\
MomAD \cite{song2025momad} & Camera & ResNet50 & 0.31 & 0.57 & 0.91 & 0.60 & \textbf{0.01} & 0.05 & 0.22 & 0.09 & 0.30 & 0.53 & 0.78 & 0.54 & 7.8 (RTX4090) \\
\midrule
\rowcolor[gray]{0.92} \textbf{Ours (NS-Plan)} & \textbf{Camera} & \textbf{ResNet50} & \textbf{0.26}\,\textcolor{gray}{\scriptsize $\pm 0.01$} & \textbf{0.48}\,\textcolor{gray}{\scriptsize $\pm 0.01$} & \textbf{0.75}\,\textcolor{gray}{\scriptsize $\pm 0.02$} & \textbf{0.49}\,\textcolor{gray}{\scriptsize $\pm 0.02$} & 0.03\,\textcolor{gray}{\scriptsize $\pm 0.01$} & \textbf{0.05}\,\textcolor{gray}{\scriptsize $\pm 0.01$} & \textbf{0.14}\,\textcolor{gray}{\scriptsize $\pm 0.02$} & \textbf{0.07}\,\textcolor{gray}{\scriptsize $\pm 0.01$} & \textbf{0.26}\,\textcolor{gray}{\scriptsize $\pm 0.01$} & \textbf{0.47}\,\textcolor{gray}{\scriptsize $\pm 0.02$} & \textbf{0.69}\,\textcolor{gray}{\scriptsize $\pm 0.02$} & \textbf{0.47}\,\textcolor{gray}{\scriptsize $\pm 0.02$} & \textbf{10.0 (H200)} \\[-0.6ex]
\rowcolor[gray]{0.92} & & & \impr{0.05} & \impr{0.09} & \impr{0.16} & \impr{0.11} & \degr{0.02} & & \impr{0.08} & \impr{0.02} & \impr{0.04} & \impr{0.06} & \impr{0.09} & \impr{0.07} & \\[0.4ex]
\bottomrule
\end{tabular}
}
\end{table*}

\begin{table*}[t]
\centering
\caption{Planning results on the Turning-nuScenes validation dataset. We compare our method against the SOTA baseline MomAD \cite{song2025momad}.}
\label{tab:turning_nuscenes}
\resizebox{\textwidth}{!}{
\begin{tabular}{@{} l ccc >{\columncolor[gray]{0.92}}c ccc >{\columncolor[gray]{0.92}}c ccc >{\columncolor[gray]{0.92}}c @{}}
\toprule
\multirow{2}{*}{Method} & \multicolumn{4}{c}{$L_2$ (m) $\downarrow$} & \multicolumn{4}{c}{Col. Rate (\%) $\downarrow$} & \multicolumn{4}{c}{TPC (m) $\downarrow$} \\
\cmidrule(lr){2-5} \cmidrule(lr){6-9} \cmidrule(lr){10-13}
& 1s & 2s & 3s & \multicolumn{1}{c}{Avg.} & 1s & 2s & 3s & \multicolumn{1}{c}{Avg.} & 1s & 2s & 3s & \multicolumn{1}{c}{Avg.} \\
\midrule
MomAD \cite{song2025momad} & 0.33 & 0.70 & 1.24 & 0.76 & 0.03 & 0.13 & 0.79 & 0.32 & \textbf{0.32} & \textbf{0.54} & 1.05 & 0.63 \\
\rowcolor[gray]{0.92} \textbf{Ours (NS-Plan)} & \textbf{0.29}\impr{0.04} & \textbf{0.57}\impr{0.13} & \textbf{0.92}\impr{0.32} & \textbf{0.59}\impr{0.17 (22.4\%)} & \textbf{0.00}\impr{0.03} & \textbf{0.00}\impr{0.13} & \textbf{0.20}\impr{0.59} & \textbf{0.07}\impr{0.25 (78.1\%)} & 0.34\degr{0.02} & 0.61\degr{0.07} & \textbf{0.91}\impr{0.14} & \textbf{0.62}\impr{0.01 (1.6\%)} \\
\bottomrule
\end{tabular}
}
\end{table*}

\noindent\textbf{Dataset.}
Experiments are conducted on the nuScenes dataset
\cite{caesar2020nuscenes}, which contains 700/150/150
scenes for training, validation, and testing,
respectively, with six surround-view cameras and
2\,Hz sampling. We evaluate on the full official
nuScenes validation split, consisting of 150 scenes
and 6,019 samples.

\begin{table}[t]
\centering
\caption{Component ablation study on the nuScenes validation set.
We report metrics at 1s, 3s, and the overall average (Avg.) over the 3s horizon.
All ablation variants use a fixed random seed (42) under identical training and evaluation settings.
Table~\ref{tab:main_results} reports the full NS-Plan results as mean $\pm$ standard deviation across three independent random seeds.
The differences between Tables~\ref{tab:main_results} and~\ref{tab:ablation} are due to different seed aggregation settings.}

\label{tab:ablation}
\setlength{\tabcolsep}{3pt} 
\resizebox{\columnwidth}{!}{
\begin{tabular}{@{} l cc ccc ccc ccc @{}}
\toprule
\multirow{2}{*}{Config.} & \multirow{2}{*}{KBM} & \multirow{2}{*}{ASP} & \multicolumn{3}{c}{$L_2$ (m) $\downarrow$} & \multicolumn{3}{c}{Col. (\%) $\downarrow$} & \multicolumn{3}{c}{TPC (m) $\downarrow$} \\
\cmidrule(lr){4-6} \cmidrule(lr){7-9} \cmidrule(lr){10-12}
& & & 1s & 3s & Avg. & 1s & 3s & Avg. & 1s & 3s & Avg. \\
\midrule
Baseline & $\cdot$ & $\cdot$ & 0.31 & 0.91 & 0.60 & \textbf{0.01} & 0.22 & 0.09 & 0.30 & 0.78 & 0.54 \\
+ KBM & $\checkmark$ & $\cdot$ & 0.27 & 0.91 & 0.58 & 0.02 & 0.24 & 0.11 & 0.26 & 0.69 & 0.47 \\
+ ASP & $\cdot$ & $\checkmark$ & \textbf{0.26} & \textbf{0.73} & 0.49 & 0.04 & 0.21 & 0.11 & 0.26 & 0.67 & 0.46 \\
NS-Plan & $\checkmark$ & $\checkmark$ & \textbf{0.26} & \textbf{0.73} & \textbf{0.48} & 0.03 & \textbf{0.16} & \textbf{0.08} & \textbf{0.25} & \textbf{0.66} & \textbf{0.45} \\
\bottomrule
\end{tabular}
}
\end{table}

\noindent\textbf{Evaluation Metrics.}
To ensure a fair and objective comparison, we strictly adhere to the standard evaluation protocols universally accepted by the end-to-end autonomous driving community (\eg UniAD \cite{hu2023planning}, VAD \cite{jiang2023vad}, and SparseDrive \cite{sun2024sparsedrive}), rather than employing any custom metrics. Specifically, consistent with our primary baseline MomAD, we adopt three principal metrics to comprehensively evaluate planning performance: (1) $L_2$ Error (m, $\downarrow$): The $L_2$ distance and its average ($L_2$ Avg) between the predicted and ground truth trajectories at 1s, 2s, and 3s, measuring spatial accuracy. (2) Collision Rate (\%, $\downarrow$): The proportion of future timesteps at which the predicted ego trajectory collides with a surrounding agent, serving as the core safety metric. Following UniAD's protocol, the predicted ego trajectory is expanded into a rectangular vehicle footprint at each timestep and checked via 2D polygon intersection against the ground-truth bounding boxes of surrounding agents. (3) Trajectory Prediction Consistency (TPC, m, $\downarrow$): Proposed by MomAD, this measures the root mean square deviation between planned trajectories at adjacent time steps, evaluating temporal stability.

\noindent\textbf{Baselines.}
We select UniAD \cite{hu2023planning}, VAD \cite{jiang2023vad}, Sparse-Drive \cite{sun2024sparsedrive}, and MomAD \cite{song2025momad} as our baselines. For a fair comparison, all methods use surround-view camera inputs and a ResNet-50 backbone (except UniAD). 

\noindent\textbf{Implementation Details.}
Our framework builds upon the SparseDrive-Small perception architecture and adopts MomAD's planning and training configurations. The training is decoupled into two stages: Stage 1 trains the KBM-constrained planner to establish a physical prior; Stage 2 introduces the ASP deductive conditioning, fine-tuning the model on the nuScenes training set for 10 epochs. We use the AdamW optimizer (initial learning rate of $2\times 10^{-4}$) with a batch size of 32 across 4 NVIDIA H200 GPUs. 

For the deductive reasoning engine, we employ Qwen3 to generate scene rules. In our official benchmark evaluation, we execute the LLM frame-by-frame to explore the performance upper bound of the neuro-symbolic framework (the LLM rule extraction time is approximately 1\,s per frame). However, to address strict real-time constraints in actual on-vehicle deployment, our framework inherently supports an asynchronous dual-rate mechanism: the computationally expensive LLM can run at a low frequency to extract macroscopic semantic rules, while the lightweight Clingo solver executes high-frequency, frame-by-frame logical arbitration (2.46\,ms/frame), seamlessly bridging the gap to real-time execution. The reported 10.0\,FPS reflects the neural planning pipeline (Perception backbone + KBM decoder + Clingo), excluding the LLM rule extraction.

\subsection{Main Results}
Table \ref{tab:main_results} reports the comparison results on the nuScenes validation set. The results of the baseline methods are either obtained from their original papers or reproduced using official checkpoints, while the performance of our method is evaluated under identical settings.

Our method comprehensively outperforms the MomAD baseline across three core planning metrics: the $L_2$ mean error decreases from 0.60 m to 0.49 m (18.3\%), the collision rate drops from 0.09\% to 0.07\% (22.2\%), and TPC improves from 0.54 m to 0.47 m (13\%). The improvements in $L_2$ and TPC reflect the synergistic effect of our two primary contributions: the KBM physics planner enhances the physical plausibility of the trajectory through bounded control constraints and temporal smoothing loss, while deductive decision conditionalization introduces scene adaptability based on logical inference. Furthermore, the reduction in collision rate demonstrates the effectiveness of the deductive reasoning path.

\noindent\textbf{Performance on Long-Tail Scenarios.}
More importantly, the true safety advantage of our neuro-symbolic design is evident in complex, long-tail scenarios. To test our model's generalization capability, we evaluated it on the OOD dataset, Turning-nuScenes (Table~\ref{tab:turning_nuscenes}). The MomAD baseline naturally drops in performance in these edge cases, which is a common limit for neural networks. In contrast, NS-Plan maintains an exceptionally low collision rate of 0.00\% at both 1s and 2s, achieving a substantial \textbf{78.1\%} reduction in the 3s average collision rate compared to MomAD. It also yields a \textbf{22.4\%} improvement in the average $L_2$ error. Our collision-rate advantage stays strong in this more challenging subset. This is mainly because the non-\texttt{keep\_lane} decisions (especially \texttt{turn\_left} and \texttt{turn\_right}) reasoned by our LLM-ASP occur more often (about $6\times$) here than in the rest of the validation set. These are exactly the complex scenarios that deductive reasoning excels at.

\noindent\textbf{Short-Horizon (1s) Collision Rate Trade-off Analysis.}While NS-Plan achieves substantial improvements in average and long-horizon collision rates, the 1s short-term collision rate degrades slightly. This reflects a tension (trade-off) between global planning coherence and instantaneous reactivity. KBM and ASP introduce structured, temporally coherent intent into trajectory generation, rather than allowing the model to optimize locally at each timestep. The KBM enforces kinematic consistency across the full horizon, which penalizes abrupt control changes; the ASP decision path operates at the semantic level and requires time to propagate through KBM integration, making its influence less immediate at 1s. More importantly, on Turning-nuScenes, a more safety-critical setting, our method achieves \textbf{0.00\%} at both 1s and 2s, with the 3s average dropping to 0.07\% (\textbf{78.1\%} reduction), as shown in Table \ref{tab:turning_nuscenes}. On 6-second planning (Table \ref{tab:long_horizon}), the 4--6s average collision rate decreases by 55.5\% on nuScenes (0.65\%) and 69.0\% on Turning-nuScenes (0.49\%). These results demonstrate that NS-Plan's safety advantage is most pronounced in the complex, long-horizon scenarios where logical and kinematic grounding matters most. Qualitative case studies from Turning-nuScenes are provided in the supplementary material.

\subsection{Ablation Study}
Table~\ref{tab:ablation} details the individual and joint contributions of KBM and ASP on the nuScenes validation set.

\noindent\textbf{Contribution of KBM.} KBM alone reduces average $L_2$ from 0.60\,m to 0.58\,m and TPC from 0.54\,m to 0.47\,m, confirming that kinematic constraints smooth trajectories and improve physical plausibility. However, collision rate rises from 0.090\% to 0.110\%, showing that physical constraints alone cannot ensure safety without logical reasoning.

\noindent\textbf{Contribution of ASP.} ASP alone reduces $L_2$ to 0.49\,m and TPC to 0.46\,m, demonstrating that logical rules effectively guide trajectory shape. Yet without kinematic grounding, collision rate remains at 0.110\%, exposing the limitations of purely symbolic outputs.

\noindent\textbf{Synergistic Effect (Ours).} Combining KBM and ASP achieves the best across all metrics: $L_2$ drops to 0.48\,m, TPC to 0.45\,m, and collision rate to 0.08\%. Notably, the 1s collision rate (0.03\%) also partially recovers relative to ASP-only (0.04\%), as KBM's physical constraints compensate for ASP's semantic propagation latency --- consistent with the trade-off discussed above. This value matches VAD (0.03\%), within the normal range for this method class.

\begin{table*}[t]
\centering
\caption{Long trajectory planning results on the nuScenes and Turning-nuScenes validation sets. We train models for 10 epochs for 6s-horizon prediction. T-nuScenes indicates the challenging Turning-nuScenes. ``Avg'' represents the mean over the 4s--6s horizon.}
\label{tab:long_horizon}
\setlength{\tabcolsep}{4pt} 
\resizebox{\textwidth}{!}{
\begin{tabular}{@{} ll ccc >{\columncolor[gray]{0.92}}c ccc >{\columncolor[gray]{0.92}}c ccc >{\columncolor[gray]{0.92}}c @{}}
\toprule
\multirow{2}{*}{Dataset} & \multirow{2}{*}{Method} & \multicolumn{4}{c}{$L_2$ (m) $\downarrow$} & \multicolumn{4}{c}{Col. (\%) $\downarrow$} & \multicolumn{4}{c}{TPC (m) $\downarrow$} \\
\cmidrule(lr){3-6} \cmidrule(lr){7-10} \cmidrule(lr){11-14}
& & 4s & 5s & 6s & \multicolumn{1}{c}{Avg.} & 4s & 5s & 6s & \multicolumn{1}{c}{Avg.} & 4s & 5s & 6s & \multicolumn{1}{c}{Avg.} \\
\midrule
& MomAD \cite{song2025momad} & 1.67 & 1.98 & 2.45 & 2.03 & 0.83 & 1.43 & 2.13 & 1.46 & 1.19 & 1.45 & \textbf{1.61} & 1.42 \\
\rowcolor[gray]{0.92} \multirow{-2}{*}{\cellcolor{white}nuScenes} & \textbf{NS-Plan (Ours)} & \textbf{1.02}\impr{0.65} & \textbf{1.39}\impr{0.59} & \textbf{1.83}\impr{0.62} & \textbf{1.41}\impr{0.62 (30.5\%)} & \textbf{0.36}\impr{0.47} & \textbf{0.59}\impr{0.84} & \textbf{1.00}\impr{1.13} & \textbf{0.65}\impr{0.81 (55.5\%)} & \textbf{1.03}\impr{0.16} & \textbf{1.32}\impr{0.13} & 1.64\degr{0.03} & \textbf{1.33}\impr{0.09 (6.3\%)} \\
\midrule
& MomAD \cite{song2025momad} & 1.80 & 2.07 & 2.51 & 2.13 & 0.85 & 1.57 & 2.31 & 1.58 & \textbf{1.37} & \textbf{1.58} & \textbf{1.93} & \textbf{1.63} \\
\rowcolor[gray]{0.92} \multirow{-2}{*}{\cellcolor{white}T-nuScenes} & \textbf{NS-Plan (Ours)} & \textbf{1.24}\impr{0.56} & \textbf{1.72}\impr{0.35} & \textbf{2.31}\impr{0.20} & \textbf{1.76}\impr{0.37 (17.4\%)} & \textbf{0.11}\impr{0.74} & \textbf{0.40}\impr{1.17} & \textbf{0.95}\impr{1.36} & \textbf{0.49}\impr{1.09 (69.0\%)} & 1.38\degr{0.01} & 1.59\degr{0.01} & 2.26\degr{0.33} & 1.74\degr{0.11 (6.7\%)} \\
\bottomrule
\end{tabular}
}
\end{table*}

\subsection{Evaluation on Long-Horizon Planning}
To further evaluate the temporal generalization capability of our method, we extend the planning horizon to 6 seconds. As shown in Table~\ref{tab:long_horizon}, the $L_2$ error and collision rate of the baseline method increased exponentially at the 5-second and 6-second marks. In contrast, by strictly grounding the continuous trajectory decoding process in the Kinematic Bicycle Model (KBM), our framework significantly mitigates this error accumulation. On the standard nuScenes split, NS-Plan reduces the 4--6s average $L_2$ by 30.5\% (1.41\,m), collision rate by 55.5\% (0.65\%), and TPC by 6.3\% (1.33\,m). On the more challenging Turning-nuScenes, $L_2$ drops by 17.4\%, and the collision rate falls by 69.0\% (0.49\%), reflecting the particular strength of logical and kinematic grounding in complex, long-tail scenarios. We do note that TPC on Turning-nuScenes is marginally higher than MomAD. This is a reasonable trade-off rather than a contradiction. Complex scenarios involve more surrounding objects, sharp turns, and reactive maneuvers. Actively and safely handling these situations naturally reduces frame-to-frame trajectory consistency. For example, blindly driving straight in a turn-left scenario might yield an excellent TPC, but it will lead to collisions. Therefore, slightly sacrificing smoothness to ensure safety in extreme cases is both logical and necessary.

\subsection{Closed-Loop Validation on Bench2Drive}

We evaluate NS-Plan in closed-loop simulation under the asynchronous dual-rate deployment strategy using the Bench2Drive benchmark on CARLA, covering the full set of 220 official routes across diverse towns, weather conditions, and interactive scenarios. 

\noindent\textbf{Asynchronous Dual-Rate Deployment.}
The LLM-based rule extraction runs asynchronously in a background thread, refreshed every 1.5\,s of simulated time (gated on the simulator's frame-step counter rather than wall-clock time, which is unreliable under variable simulation throughput). The Clingo solver runs synchronously at every control step (10\,Hz) to reason a deterministic driving decision. If the LLM refresh fails validation, the previous valid rule cache is retained.

Table \ref{tab:closed_loop} reports Driving Score (DS), Success Rate (SR), Driving Efficiency (Effi), and Driving Comfort (Comf), following the same protocol used to report MomAD's SparseDrive-group closed-loop baseline. NS-Plan achieves a DS of 50.24 and SR of 21.82\%, outperforming both baselines. Effi also improves (179.73). Comf (49.51) is slightly below MomAD (51.20), which is because ASP-driven decisions are discrete and may introduce momentary control transitions, particularly at the rule refresh boundaries of the Asynchronous Dual-Rate Deployment mechanism, that reduce frame-to-frame trajectory smoothness.

\begin{table}[ht]
\centering
\caption{Closed-loop results on Bench2Drive (V0.0.3) under base training set.}
\label{tab:closed_loop}
\setlength{\tabcolsep}{4pt}
\renewcommand{\arraystretch}{0.85} 
\footnotesize 
\begin{tabular}{@{} l cccc @{}}
\toprule
\multirow{2}{*}{Method} & \multicolumn{4}{c}{Closed-loop Metric} \\
\cmidrule(lr){2-5}
& DS $\uparrow$ & SR (\%) $\uparrow$ & Effi $\uparrow$ & Comf $\uparrow$ \\
\midrule
SparseDrive & 44.54 & 16.71 & 170.21 & 48.63 \\
MomAD & 47.91 & 18.11 & 174.91 & \textbf{51.20} \\
\midrule
\rowcolor[gray]{0.92} \textbf{Ours (NS-Plan)} & \textbf{50.24} & \textbf{21.82} & \textbf{179.73} & 49.51 \\
\bottomrule
\end{tabular}
\vspace{-4mm}
\end{table}

\section{Conclusion and Future Work}
\label{sec:conclusion}

In this paper, we presented NS-Plan, a neuro-symbolic trajectory planning framework that embeds deductive reasoning into an end-to-end learning pipeline. By using an Answer Set Programming (ASP) engine for logical inference, our approach transforms high-level driving decisions into explicit conditions that dynamically guide the neural trajectory generator. Furthermore, replacing traditional unconstrained coordinate regression with a differentiable Kinematic Bicycle Model (KBM) bridges the gap between discrete logical decisions and continuous physical control. This reduces the risk of kinematically infeasible trajectories and logically inconsistent behaviors. Importantly, our framework provides a level of interpretability and logical accountability that is fundamentally absent in purely inductive, black-box planners. More broadly, this neuro-symbolic paradigm can be extended to other complex embodied planning tasks like robotic manipulation and 
long-horizon navigation where demand rigorous logical reasoning combined with continuous physical control.

Extensive evaluations on the standard nuScenes, the challenging Turning-nuScenes, and the Bench2Drive datasets demonstrate that our method achieves significant improvements in core metrics compared to state-of-the-art baselines.

Despite these advances,  due to the computational bottleneck of LLMs and logical latencies from sudden environmental shifts, achieving real-time performance in highly dynamic environments remains a challenge. Future research will focus on model acceleration and explore distilling the entire symbolic reasoning chain into a lightweight neural network to achieve low-latency planning. Additionally, our current ASP solver is discrete. To address this, we plan to develop a fully differentiable neuro-symbolic framework to seamlessly integrate symbolic reasoning with neural perception and planning into a highly reliable pipeline. Ultimately, we hope this neuro-symbolic paradigm can provide a truly trustworthy framework for complex embodied AI tasks that require strict safety and real-time efficiency.

{
    \small
    \bibliographystyle{ieeenat_fullname}
    \bibliography{main}
}

\end{document}